\documentclass{article}

\usepackage{arxiv}

\usepackage{natbib}
\usepackage[utf8]{inputenc} 
\usepackage[T1]{fontenc}    
\usepackage{hyperref}       
\usepackage{url}            
\usepackage{booktabs}       
\usepackage{amsfonts}       
\usepackage{nicefrac}       
\usepackage{microtype}      
\usepackage{lipsum}
\usepackage{graphicx}
\usepackage{bm}
\usepackage{enumerate}
\usepackage[flushleft]{threeparttable}
\usepackage{multirow}
\usepackage{amsthm,amsmath}

\title{Subsampling Winner Algorithm for Feature \\ Selection in Large Regression Data}

\author{
 Yiying Fan \\
  Department of Mathematics and Statistics\\
  Cleveland State University\\
  2121 Euclid Avenue\\
  Cleveland, OH 44115-2214 \\
  \texttt{y.fan67@csuohio.edu} \\
   \And
 Jiayang Sun \\
  Department of Statistics\\
  George Mason University\\
  4400 University Drive\\
  Fairfax, VA 22030\\
  \texttt{jsun21@gmu.edu} \\
}

\begin{document}
\maketitle
\begin{abstract}
Feature selection from a large number of covariates (aka features) in a
regression analysis remains a challenge in data science, especially in terms of its potential of scaling to ever-enlarging data and finding a group of scientifically meaningful features.
 For example,  to develop new, responsive drug targets for ovarian cancer,    the actual false discovery rate (FDR) of a practical feature selection procedure must also match  the target  FDR.
The popular approach to feature selection, when true features are sparse,
is to use a penalized likelihood or a shrinkage estimation, such as
a LASSO, SCAD, Elastic Net, or MCP procedure (call them {\em benchmark} procedures). We present a different approach using a new subsampling method, called the Subsampling Winner algorithm (SWA).   The central idea of SWA is analogous to that used for the selection of US national merit scholars. SWA uses a `base procedure' to analyze each of the subsamples, computes the scores of all features according to the performance of each feature from all subsample analyses, obtains the `semifinalist' based on the resulting scores, and then determines the `finalists,' i.e., the most important features. Due to its subsampling nature, SWA can scale to data of
any dimension in principle. The SWA also has the best-controlled actual  FDR in comparison with the benchmark
procedures and the randomForest, while having a competitive true-feature discovery rate.  We also suggest  practical add-on strategies to SWA with or without a penalized benchmark procedure  to further assure the chance of `true' discovery.  Our application of  SWA
to the ovarian serous cystadenocarcinoma specimens from the Broad Institute revealed functionally important genes and pathways, which we verified by additional genomics tools. This second-stage investigation is essential in the current discussion of the proper use of  P-values.
\end{abstract}


\section{Introduction}

Feature  selection is important in data mining and knowledge
discovery. It has many applications, including 1) finding and
ranking important disease genes; 2) reducing nuisance features
that mask truth or terminate a computation due to the size or
memory limitation of a current computing software architecture
[\citet{liu}]; 3) building a parsimonious model or performing
a dimension reduction; and 4) improving efficiency in storage,
communication, or computation.

    A {\em unified view} for feature selection in supervised
learning is that finding important features is equivalent to
an optimization problem that seeks a solution  ``$\bf a$''
that maximizes an objective function $J({\bf a}|{\bf D})$ over
a domain of decisions ``$\bf a$'', given a data set $\bf D$.
Here ${\bf D}={\bf (X, y)}$, where $\bf X$ is a $n\times p$
matrix representing measurements of $p$ covariates measured
on $n$ subjects, and $\bf y$ denotes the outcomes of these
$n$ subjects. The number $p_0$ of true important features
(defined explicitly below in Section 2) from the $p$
covariates is unknown and typically sparse. $J$ depends on
a statistical {\em objective} and a suitable {\em model} for
analyzing the data $\bf D$. The value of ${\bf a}$, which maximizes
$J$ is indicative of the importance of features. For example, $J$
is a penalized summation of squares of ``residuals'' in a linear
regression context, or a measure of predictive classification
rate  in a classification model.

With a large $p$, seeking an exact optimization of $J$ may be
impossible or inefficient.  Approximate  solutions or clever
alternatives are often necessary for practical use. These
alternatives include Local Quadratic Approximation (LQA) for
minimizing a penalized likelihood with a SCAD penalty
[\citet{fan}], LAR [\citet{lar}] for speeding up LASSO
[\citet{tib}], MCP [\citet{zhang}] that advocates using a
minimax concave  penalty, and iterative procedures, such as
stepwise procedure/ELSA [\citet{kim}],  generalized bagging
[\citet{Breiman96}] and boosting [\citet{Schapire}] procedures
for tree based models. As a current state-of-the-art technique,
the Elastic Net [\citet{learn}] is a regularized regression
method that linearly combines the L1 and L2 penalties of the
LASSO and ridge methods. This procedure and its cousins, SCAD
and MCP, all belong to the class of penalized (likelihood)
strategies. They have been studied extensively to generalize
for example, to a smooth regression model that subjects to
different covariate effects within a group by \citet{guo};
for their asymptotic properties by \citet{yu};  or for asymptotic
optimality by \citet{zhangl}; for multivariate regression
modeling with a large number of responses by \citet{sun}; and
for confidence intervals in high-dimensional linear regression
[\citet{cai17}]. The Random Forest [\citet{Breiman01}] is the
benchmark of tree-based methods for producing importance of
predictors (or features) in the application of ensemble learning
and regression. These sophisticated procedures assume that
target data are within the size and memory limitation of these
procedures' software implementation. The ever-increasing volume
of data continuously challenges this assumption. When a
limit is reached, one strategy is to use a sample of the data.
However, large data are often heterogenous, a subsample or even
a few subsamples may not adequately represent the full data set.
Another strategy is to use a newer machine with a larger space
and memory, and a new software implementation. On the other hand,
when the true number of features is only a few, inclusion of all
features, even only a few hundreds or thousands into a (standard)
statistical analysis procedure can mask true features. See
\citet{liu2} for subsampling for PCA, an unsupervised learning.
The third  approach is using a rough dimension reduction
methodology or domain knowledge to prescreen the features. Here,
we consider a different approach, developing {\em an ensemble that
uses an adequate number  of multiple subsamples and
the outcomes from these subsamples  strategically}. We also
wish to require  this ensemble procedure {\em simple to implement} and
{\em easy to scale} to ever-enlarging dimension or size of data.
Our Subsampling Winner Algorithm (SWA) is developed based on
thess ensemble requirements. Our SWA differs from the
``subsampling procedures" in \citet{wang}, which subsample from
$n$ observations; SWA subsamples from $p$ variables or features,
and hence the traditional statistics based on a large number $n$ of observations
does not apply here. SWA bears some similarity to the philosophy
behind the bagging and boosting for a tree based model in selecting
the best split at each node of the tree. SWA applies simple, standard
procedures on subsamples strategically as \citet{donoho} did for
Higher Criticism (HC) motivated by a Tukey's idea, and hence SWA
can be used for large data without adding too much complexity in
its algorithm. By its subsampling nature, SWA reduces not only
the requirement of the memory or size, but also the complexity
that arises from many masking nuisance features in the whole sample.

This paper focuses on developing a practical SWA  for feature
selection in linear regression from large data. In Section 2,
our sparse linear model is provided with an ovarian cancer data
that motivated our development of SWA for regression. In Section 3, the Subsampling Winner Algorithm (SWA)
is introduced. In Section 4, the procedures for specifying the
SWA parameters are provided. In Section 5, the SWA's parameter
specification procedure is validated, and  a  comprehensive
comparison of SWA performance with those of  benchmark procedures
are given. The benchmark procedures  include the Elastic Net,
SCAD, MCP, and Random Forest. In Section 6, the SWA is applied to
ovarian cancer data, where we also provided a ``double assurance''
procedure for practical applications. We found that SWA has
excellent, automatic control of false discovery rates (FDR), and is competitive
in capturing true features, i.e., with a good true discovery rate
(TDR) when the FDRs are {\em equalized} for all comparison procedures.
Also, the features discovered  by SWA from  the ovarian cancer data
contain functionally important genes and pathways, setting a basis
for future research into much needed development of drug targets for ovarian cancer. The article
is concluded with a discussion, summary, and recommendations
in Section 7, including an analogy of SWA with
the deep learning principle.
A simple proof of our proposition is given in the
appendix.

\section{Model and  Data}
Consider data ${\bf (X,y)}$ from a linear model
\begin{equation}
\bf{y} =  \bf{X} \bm{\beta}  + \bm{\varepsilon} ,\quad\quad \bm{\varepsilon} \sim N(0,\sigma^2I)
\label{eq:model}
\end{equation}
relating the outcome $\bf{y}$ and its $p$-dimensional covariates
(aka features) via their measurements in ${\bf X}_{n\times p}$,
and the unknown parameter vector $\bm\beta=(\beta_{1},\cdots,$ $
\beta_{p})^T$. Here, the n-dimensional error vector
$\bm{\varepsilon}$ is assumed to come from a homoscedastic normal
distribution with mean $0$ and variance $\sigma^2$, for simplicity.
This simple model (\ref{eq:model}) is adequate for many data
applications, though the data may need to be preprocessed or
transformed before its modeling by (\ref{eq:model}), such as the motivating data
below. This model is also the most basic model for evaluating
a paradigm or direction change from a status quo  such as \citet{cai17} did for
confidence intervals in high dimension. A generalization of
(\ref{eq:model}) to more general error structures will be discussed
in section 7. The challenge here, as it is in many recent modern
procedures, is when $p \gg n$. These recent procedures have been
based on the exploitation of sparsity assumption of the large
$\bm\beta$ vector, which assumes that only a small number $p_0$
out of $p$ components of $\bm\beta$ are non-zero.  We assume the
same moderate sparsity with $p_0 \ll\frac{n}{log p}$ as \citet{cai17}
did. Our objective here is to identify the set of `true features'
that are associated with the `covariates' corresponding to non-zero
coefficients $\beta$'s.

{\bf Data}. In human cancer genomic research, feature selection
plays an important role in analyzing big data. Messenger RNA (mRNA)
are RNA molecules that transmit genetic information from DNA to the
ribosome, where they control the amino acid sequence of gene expression.
In particular, for the ovarian cancer data from the \citet{broad},
we are interested in the identification of mRNAs that are directly linked
to an important gene expression (aka, the response ${\bf y}$) that has
a significant impact on the overall survival of ovarian cancer patients.
Analyzing this publicly
available data set is in response to the NIH call for ``{\em
Secondary Analysis and Integration of Existing Data to Elucidate
the Genetic Architecture of Cancer Risk and Related Outcomes (R01)}''
(\url{https://grants.nih.gov/grants/guide/pa-files/PA-17-239.html}).

The {\bf important}  mRNA genes would influence the response and
have non-zero $\beta$'s in explaining the response in (\ref{eq:model}).
The ovarian cancer data contain $p=12042$ mRNA gene expression profiles
derived from $n=561$ ovarian serous cystadenocarcinoma specimens
available from the Broad Institute's GDAC. The data have been
normalized and analyzed for an association study \citep{broad}. Hence,
Equation (\ref{eq:model}) is an adequate approximate model for our
application to this normalized dataset.

Clearly $p\gg n$ in this ovarian cancer data. Thus, an adhoc
or a standard statistical feature selection procedure is inconsistent.
An adequate procedure designed specially for large-$p$-small-$n$
problems is needed. SWA described below is suitable for the case
of $p\gg n$ and will be applied to this ovarian cancer data to find
significant genes linked to an important response ${\bf y}$. The
resulting ``important'' genes will help to validate candidate target
genes in literature, or mine further important features that were not
presented in the existing biological literatures.

\section{Subsampling Winner Algorithm}

This section presents our Subsampling Winner Algorithm ({\bf SWA})
for linear regression with data from equation (\ref{eq:model})
when $p \gg n$. Given a suitable base procedure {\bf h} and a
scoring algorithm {\bf w} that ranks features, our SWA goes
through 4 steps: (1) SWA subsamples data and performs subsample
analysis using a base procedure to each of $m$ subsamples of
size $s$, for $s<n$; (2) SWA ranks the subsample models and features;
(3) SWA obtains $q$ semifinalists from the ranks of features;
(4) SWA analyzes the $q$ semifinalists to capture the finalists,
i.e., the most important features using the base, or an enhanced base
procedure. Hence, an SWA depends on $s,m,q$ and {\bf h} and {\bf w}:
\begin{equation}
    SWA(s,m,q | {~\bf h,w}),
\label{eq:model1}
\end{equation}
where {\bf h,~w} are the base and scoring functions, respectively.

For the linear regression model (\ref{eq:model}), {\bf h} can be
simply the standard least square procedure for subsamples of $s$
dimension of $n$ data points ($s<n$). We define $\bf w$ based on
the scores of features in each sub-model weighed by the sub-model
performance as given in the score function (\ref{eq:weight}) below.

{\bf Algorithm 1: SWA for Regression in Large Data}
\begin{enumerate}
    \item \textbf{Subsample analysis.} For each $i=1,\ldots,m$, \{
        \begin{enumerate}
            \item \textbf{Sample candidate features.}
                Randomly draw $s$ sub-columns from $p$ columns of
                ${\bf X}_{n\times p}$ to obtain an $s$ dimensional
                sub-matrix, still denoted as ${\bf X}_{n\times s}$
                for simplicity though having abused the notation.
            \item \textbf{Least squared regression.}
                Fit ${\bf y}\sim {\bf X}_{n\times s}$ by {\bf h}, the standard
                least square procedure (and allowing an additional stepwise variable
                selection on this resulting fit, as a user specified option).
            \item \textbf{Grade candidates locally.}
                Store the resulting `Residual Sum of Squares' and
                estimated `$t$ statistic' of $\beta's$ as $RSS_i$ and $t_{ij}$,
                for $j=1,\ldots,p$, where the $t_{ij}$ for the
                unsampled features ${\bf X}_{n\times s}^{(-)}$ are set to zero.\\
                \}
        \end{enumerate}
    \item \textbf{Rank sub-models.}
        Rank $RSS_i$ for $i=1,\ldots,m$, and let $RSS_{(i)}$ be the
        $s$ smallest Residual Sum of Squares and $t_{(i)j}$ be the
        corresponding t-values of $\beta$, for $i=1,\ldots,s$, $j=1,\ldots,p$.

    \item \textbf{Rank all individuals to obtain semi-finalists.}
        Compute the scores of all features for $j=1,\ldots,p$,
        by the {\em scoring function}:
        \begin{equation}
            w_j=\frac{1}{S_j}\sum_{i=1}^s \frac{1}{\sqrt{RSS_{(i)}}}|t_{(i)j}|,
            \quad\quad {\rm where~} S_j=\sum_{i=1}^s I\{t_{(i)j}\neq0\}.
        \label{eq:weight}
        \end{equation}
        Retain the top $q$ features that correspond to the $q$ largest
        $w$ values, as the {\em semifinalists}.

    \item \textbf{Select finalists.}
        Fit a linear regression of $\bf y$ on the $q$ semifinalists by
        {\bf h}. The features with p-values less than $5\%$ (after a
        multiplicity adjustment) are selected as the finalists. A
        further stepwise selection is also an option from this final
        regression fit.
\end{enumerate}

We shall evaluate the SWA by comparing SWA with its benchmarks, in terms of the actual false discovery rate and true discovery rate of important features.

 This SWA algorithm is simple. It can subsample from large $``p"$ data
to select important features with no restriction on the size of $p$.
It therefore may be called ``dimensionless'' and will be shown to
be competitive  in Section 5.

\section{Procedures for SWA's Parameter Selection}
This section addresses the specification of
$m$, $s$ and $q$ in (\ref{eq:model1}) for the SWA in regression.
We use SWA$(s,m,q)$ in stead of SWA$(s,m,q | {~\bf h,w})$ hereafter.

\subsection{\bf The Choice of $s$}

Denote $F_{true}$ to be the set of $p_0$ true features and ${F}_{SW}$
to be the set of features selected by SWA.  If the fixed $s\geq p_0$,
then as given by  {\bf Proposition 1} below, a chance that all $p_0$
features are selected in one subsample is close to $1$ when $m$ is
large. The estimated coefficients  from this subsample analysis would
then be consistent for a fixed $s<n$, as   $n\rightarrow \infty$ [\citet{rao}].

 In general, we assume that data come from an
{\em identifiable model}  in the sense that (1) features are {\em sparse}
with  $p_0\ll n/ \log(p)$ [\citet{cai18}]; (2) models with more true
features are {\em dominant}   than those with less true features, leading to small RSS; and
(3)  the  {\em SOIL condition}  by \citet{ye} is satisfied as
$m\rightarrow \infty$. See more discussion on this SOIL condition below.
Hence, we would then have that if $s\ge p_0$,

\begin{equation}
    P({F}_{SW} \Delta F_{true})\rightarrow 0,
    \quad\quad {\rm as~} m \rightarrow \infty {\rm ~ and~} n\rightarrow \infty.
\end{equation}
In other words, when $s\geq p_0$, the features selected by the SWA have
a large probability to be consistent to the set of true nonzero
features when the model is identifiable. However, the true number of
features $p_0$ is unknown. If $s$ is too small, we may not be able
to capture with a high probability all the important features in any
subsample. On the other hand, if $s$ is too large, a large $s$ can
slow down computation dramatically and may mask important features.
 Hence $P({F}_{SW} \Delta F_{true})\rightarrow 0$ as $n\rightarrow \infty$,
where ``$\Delta$'' denotes the symmetric difference of two sets
$F_{SW}$ and $F_{true}$.

{\bf Rule of Thumb.}
Our simulation and experience recommend a rule of thumb in selecting $s$:
\begin{equation}
    p_0\leq s \leq  3p_0.
\end{equation}
This rule of thumb does not require
knowing $p_0$ but some range of $p_0$. When the range is still quite
uncertain, we propose a multipanel diagnostic plot for choosing $s$,
constructed based on the following algorithm. \\

{\bf Algorithm 2: Multipanel Plot Algorithm (MPA) for selecting $s$}
\begin{enumerate}
    \item \textbf{Initialize a set of $s$ values}:
        $S=\{s_i:i=1,\cdots,I\}$, e.g., $S=\{3,5,10\cdots\}$.
    \\
    \item \textbf{Arrange scree plots of feature weights for each of the s values in I panels}.
        Run the SWA$(s_i,m,q)$ for $i=1,\cdots,I$. Name the resulting weights
        $w_j$ in (3) for each $i$ as $w_{ij}$, for $j=1,\cdots,p$ and $i=1,\cdots,I$.
        Plot $\{(j,w_{ij}): j=1,\cdots,p\}$ for $i=1,\cdots,I$ to obtain $I$
        simultaneous panel plots in both fixed and free-scales to examine the changes
        in both magnitude and shape of these plots. See Figures 1 \& 2.

\newpage

    \item \textbf{Identify the minimal stable point of s}.
        A value of $s$ is a stable point, if its scree plot has an obvious ``elbow'' point
        and a stable ``upper arm'' set. An ``elbow'' point is the point where a significant
        change in the slope of a scree plot has occurred, typically leading to a tapering
        flat line. An ``upper arm'' set contains the points above the ``elbow'' point, for
        example, feature set $\{1,2,3\}$ in Figures 1 and 2. An ``upper arm'' set is stable
        if indices of the points in the ``upper arm'' set are the same as those of the
        ``upper arm'' set of the next $s$ value (although set orders may differ).
        A set is “relative stable" if the intersection of this set with its adjacent upper
        arm set(s) (for adjacent s values) contains a stable subset in reference to the
        neighboring sets.
    \item \textbf{Optional step}.
        If there are no clear stable plots, but some semi-stable subplots (i.e.,
        there are no definite stable sets), enlarge the set of $S$ by adding a few
        ${s_i}'s$ in the neighborhood of semi-stable plots, return to Step 2 using
        the updated $S$.

\end{enumerate}

Both practical selection procedures of $s$ are simple to implement.
The multipanel plot has been implemented in our R package ``subsamp,''
available in cran.r-project.org.

The following are two examples of using this algorithm.

\vspace{0.1in}
\noindent
{\bf Example 1}: Consider data from the model:
\begin{equation}
    y_{i} =  2X_{i1} + 3X_{i2} + 5X_{i3}  + {\varepsilon_i},
\label{eq:simu.model1}
\end{equation}
where $\varepsilon_i$, $X_{i1},\ldots,X_{i100}$ are i.i.d.
$\mathcal{N}(0,1)$, for $i=1,\ldots,20$. Hence, $p=100$,
$p_0=3$, and $n=20$. For this example, we apply the SWA by
subsampling with a repetition size $m=5000$, and $s =
\{3, 5, 6, 9, 12, 15\}$. The resulting multipanel plots of
the weight scores of the top 40 features for each choice
of $s_i$, for $i=1,\ldots,6$, are in Figures 1 
and  2. 

The fixed-scale multipanel plot Figure 1 
shows a significant change in magnitude occurred at $s= 6$
or $s=12$. Combining with the information in the free-scale plot
Figure 2, 
we see that the plot at $s=6$ has an ``elbow'' point, while the
plot at $s=12$ does not have an obvious ``elbow'' point.

\begin{enumerate}[(a)]
    \item The upper arm set is $\{1,2,3\}$ for the plot of $s=6$ and this
        set has a relatively stable subset of $\{2,3\}$.  It is important
        to note that the coefficient of $X_1$ is not as large as these of
        $X_2$ and $X_3$. So ``1'' can hide among other features.
    \item $min(s_i)=6$, obviously.
\end{enumerate}
When $s \geq 12$, true features are gradually masked more by nuisance
features. Hence, the corresponding $s=6$ would be the recommended
subsample size.


\noindent
{\bf Example 2}: Consider data from the model
\begin{equation}
    y_i={{\bf X}_i}^{T}{\bm \beta}+{\varepsilon_i},
    \quad\quad {\rm with~} {{\bf X}_i}^T=(X_{i1},\ldots,X_{i10}),
\label{eq:simu.model2}
\end{equation}
where ${\bm \beta}=(0.1, 0.5, 1, 1.5, 2, 2.5, 3, 3.5, 4, 5)^T$,
$\varepsilon_i$ and $X_{i1},\ldots,X_{i100}$ are i.i.d.
$\mathcal{N}(0,1)$, for $i=1,\ldots,80$. This is again large
$p$ $(=100)$ small $n$ $(=80)$ data with $p_0=10$ true features.
For this example we apply SWA by subsampling with a
repetition size $m=5000$, and $s = \{5, 10, 20, 30, 40, 50 \}$.
\begin{enumerate}[(a)]
    \item As shown in the fixed scale multipanel plot in Figure 3, 
    significant changes in magnitude
        occurred at $s \geq 30$. In combination with the information
        from Figure 4, 
        we observe that $s=30, 40,
        50$ lead to plots with a stable `elbow' point.
    \item The upper arm set is $\{2,3,4,5,6,7,8,9,10\}$, which is
        also stable.
    \item It is obvious that $min(s_i)=30$. It is also important
        to note that the signal of $X_1$, $X_2$ and $X_3$ are not
        as strong as the rest of true features with that of $X_1$
        being extremely weak, especially when compared with the
        variance of $\varepsilon_i$.
\end{enumerate}

 \begin{center}
    \includegraphics[width=0.8\textwidth]{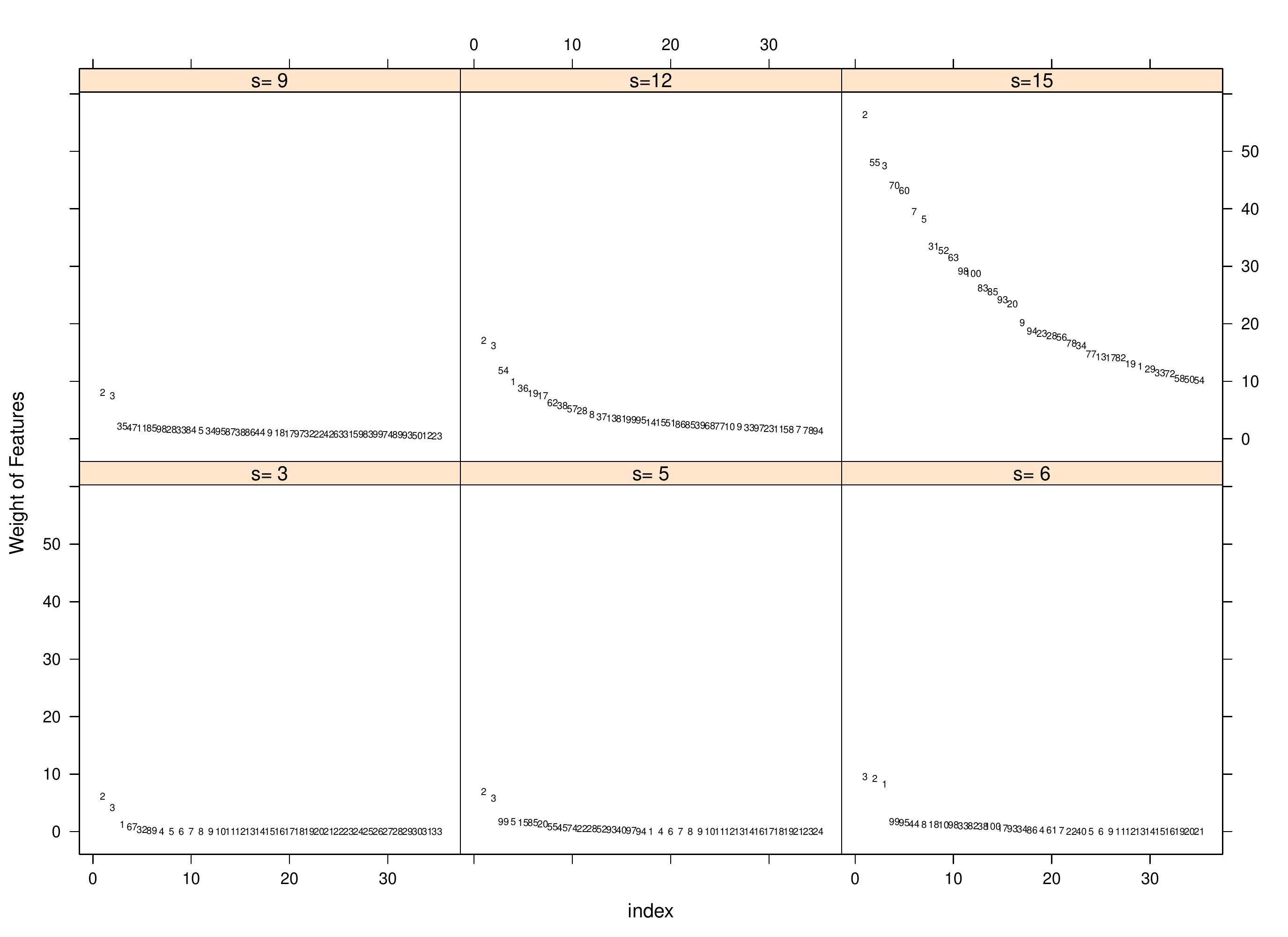}\\
\vspace{-0.02in}
    {\em Figure 1: Fixed-scale multi-panel diagnostics plot for $p_0=3$.}\\
    \includegraphics[width=0.8\textwidth]{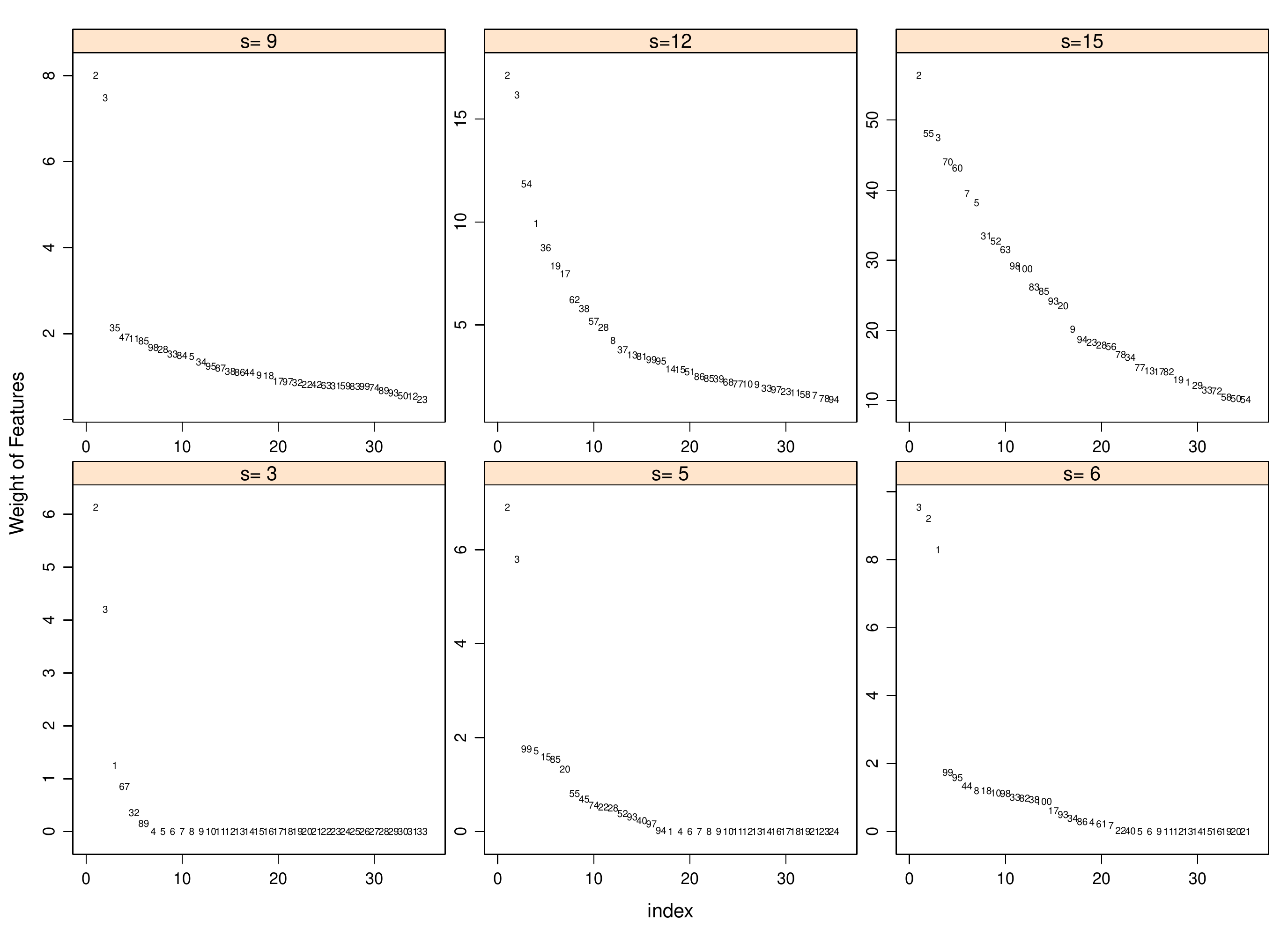}\\
    {\em Figure 2: Free-scale multi-panel diagnostics plot for $p_0=3$.}
\end{center}

\vspace{10px}
\begin{center}
    \includegraphics[width=0.8\textwidth]{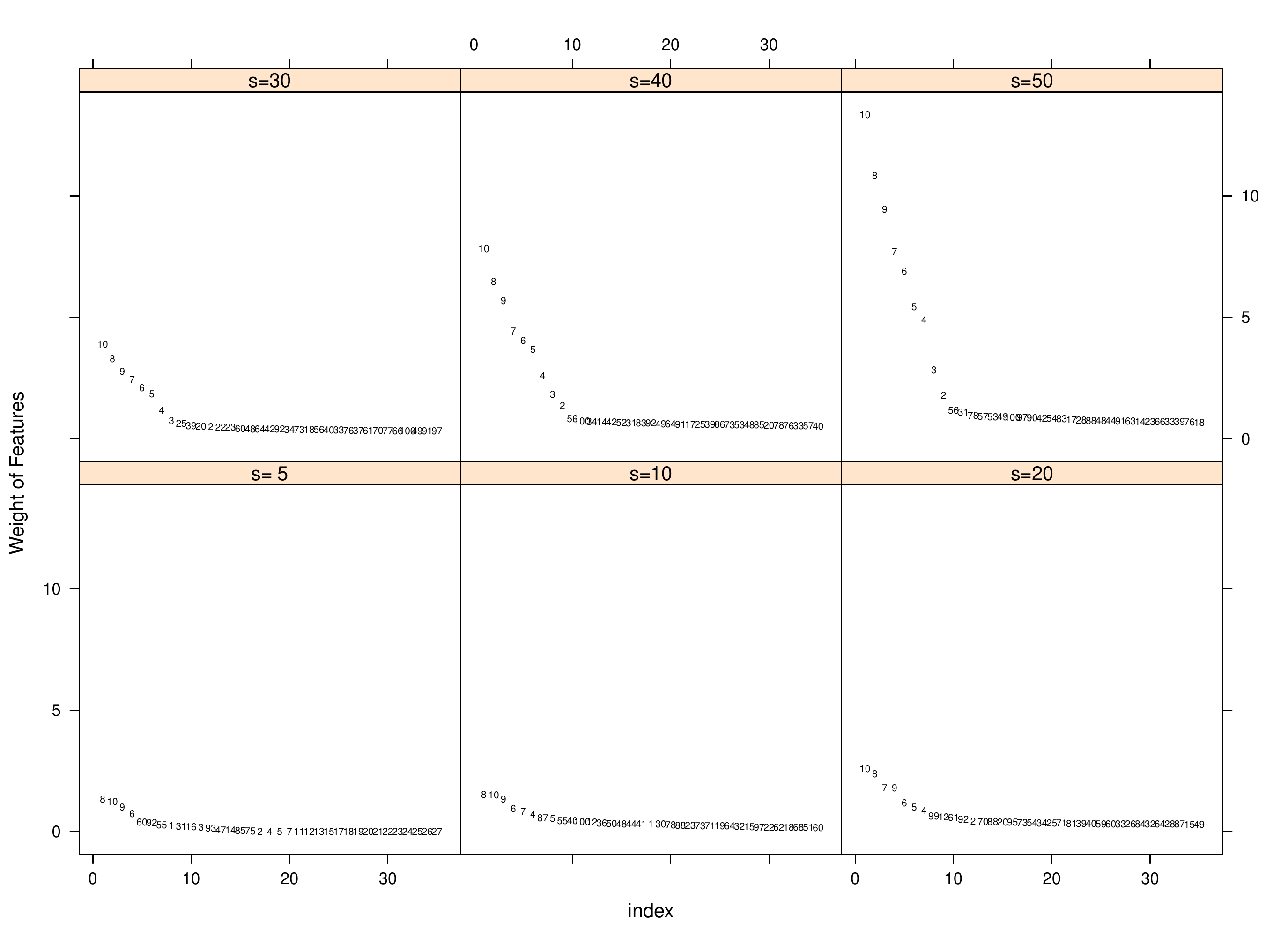}\\
  {\em Figure 3: Fixed-scale multi-panel diagnostics plot for $p_0=10$.}\\
    \vspace*{\floatsep}
    \includegraphics[width=0.8\textwidth]{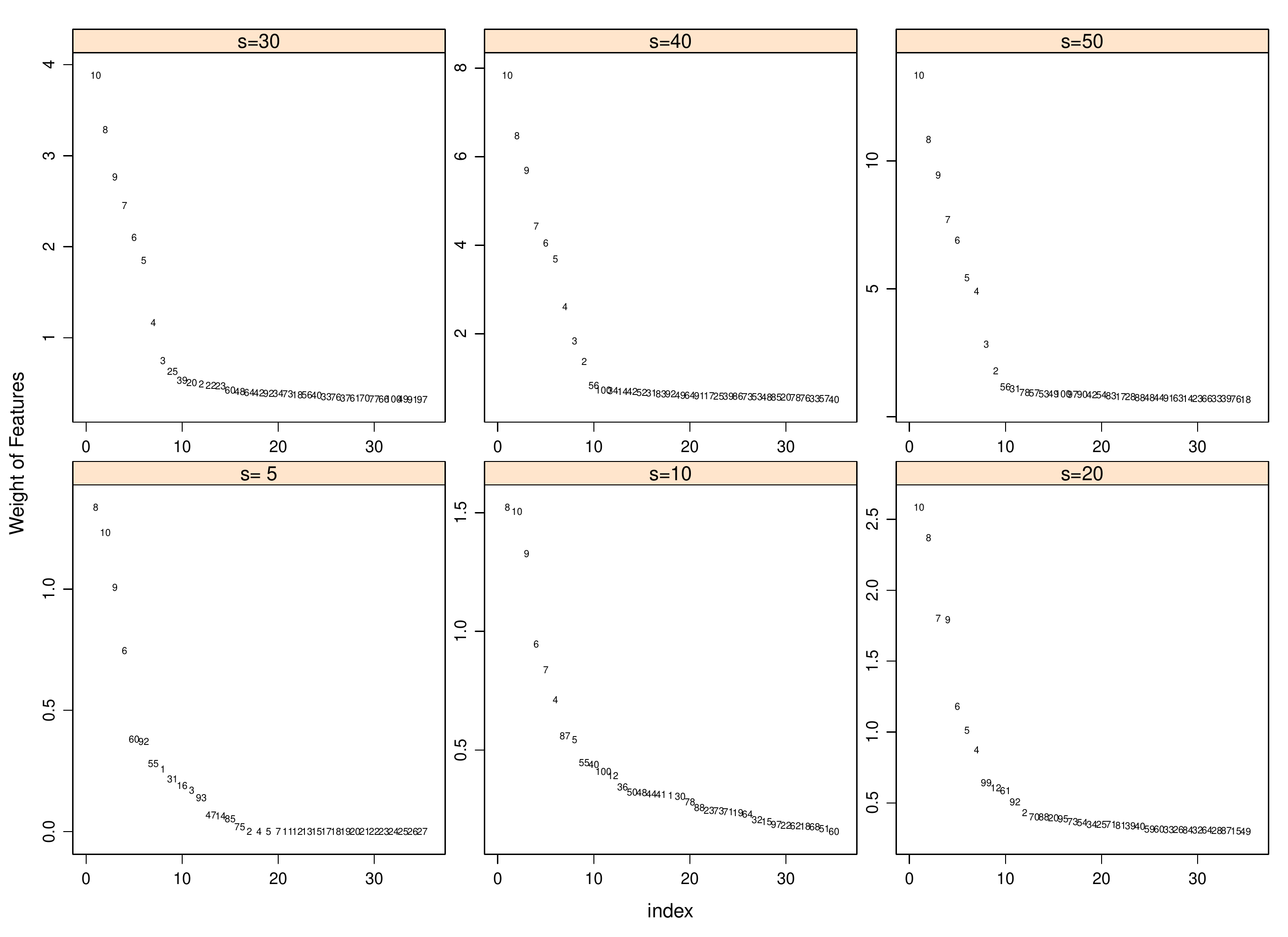}\\
 {\em Figure 4: Free-scale multi-panel diagnostics plot for $p_0=10$.}
\end{center}


\noindent
The $s=30$ is the optimal subsample size in terms of stabilization
in both scale and shape. In general, $s$ in
$p_0\leq s \leq  3p_0$ is sufficient by our experience, although
there are some cases when $s<p_0$ for a large $p_0$ may still work.
In Section 5.1, we validate systematically the multipanel procedure
for choosing s. In these two examples, we did not need to use Step
4 in Algorithm 2.

\vspace{0.03in}

{\bf Discussion of the identifiable model}.
Given the SOIL condition in the identifiable model, \citet{ye}
has shown that the ensemble model leads to a consistent feature
selection. We made the same assumption in our identifiable model.
However, in our studies, under the identifiable condition (2),
the dominance condition, the scoring functions $w_i$'s in
(\ref{eq:weight}) typically place strong-enough important
features into a semi-finalist set and hence lead to consistent
estimates of regression coefficients. This is perhaps why our
SWA procedure is competitive to the consistent penalized procedures
in our simulation studies. We conjecture that conditions (1) and
(2) in the identifiable model above would imply conditions (3),
the SOIL condition or the consistency of Random Forest
[\citet{scornet}], under some mild regularity conditions on the
strength of the signals. The ideas used to show that Adaptive
Regression by Mixing (ARM) leads to the SOIL condition [\citet{yang}]
might be generalizable to probe the consistency of our SWA procedure.
We leave the formal consistency condition and proof to future
investigation.

\subsection{\bf The Choice of $m$}

{\bf Proposition 1}. Given that $s\ge p_0$, the upper and lower bounds for
the number of independently selected subsamples of size $s$ to capture all
the important features with a probability at least $1-\gamma$ are given by

\begin{equation}
      \frac{log\gamma}{log\left(1-(\frac{s}{p})^{p_0}\right)} \le m \le   \frac{log\gamma}{log\left(1-(\frac{s-p_0+1}{p-p_0+1})^{p_0}\right)}.
 \label{eq:msize.01}
\end{equation}
A simple proof is given in the appendix.
\medskip

\noindent
{\bf Remark}:
It is worthwhile to point out that the repetition number $m$ given by
Proposition 1 is usually quite large. This is due to the consideration
of the ``worst'' case senario, in which the important features would
not be discovered unless all $p_0$ of them were selected in one of
subsamples. This ``all or none" condition is extreme.  The repetition
number needed to catch the important features is often much smaller
than the one indicated by Proposition 1. The repetition number and
its corresponding lower and upper bounds given by Proposition 1 may
be considered conservative benchmarks.

The larger $m$ is, the more computation load will be. In our R package
``subsamp,'' the default $m=5000$, has been sufficient for all examples
we tried. Parallel computing can be used to speed up the implementation
of the algorithm  process, see discussion in Section 7.

\subsection{\bf The Choice of q}
In the last step of Subsampling Winner Algorithm, in our examples and
application in Section 5 and 6, we have chosen $q=s$ in the final
(stepwise) model selection, which seems to be satisfactory for the
values of $p$ we have tried.

\section{Validation and Comparisons}
In this section, we first validate the MPA method specified in Algorithm
2, for selecting the subsample size $s$, by empirically examining the
performance of SWA with various choices of subsample size $s$ (Section 5.1).
We then compare the SWA with three benchmark procedures: Elastic
Net, SCAD and MCP, via their R code implementation ``glmnet'', ``ncvreg''
and ``plus'', and then with the ``randomForest'' procedure.
Both independent (Section 5.2) and correlated (Section 5.3) covariates
are considered in our comparative study. The ultra high dimensional case
with a prescreening procedure is also examined (Section 5.4).

\subsection{Validation of Selection Method MPA of $s$}

The model of our choice for validating the MPA is (\ref{eq:simu.model2})
given in Example 2, where $p=100$, $n=80$, $p_0=10$ with ${\bm \beta}
=(0.1, 0.5, 1, 1.5, 2, 2.5, 3, 3.5, 4, 5)^T$ representing the parameters
of true features. This model has a mixture of features including
{\em extremely weak} (with coefficients $\beta=$ 0.1 and 0.5),
{\em weak} ($\beta=$ 1 and 1.5), {\em moderate} ($\beta=$ 2 and 2.5),
{\em strong} ($\beta=3$) and {\em very strong} ($\beta=$ 3.5, 4 and 5)
features.

Using the MPA, we have determined that $s=30$ is a good choice of
subsample size for example 2. For each sample drawn from this model,
variables selected by SWA,  were recorded; p-values of the finalists
were adjusted by the simple Bonferroni method to control the
multiplicity of searching from $p=100$ covariates at $0.05$
family-wise error (FWE) rate. We repeated this random sampling and
analysis $1000$ times independently.

Tables 1 and 2 show the effects of different choices of $s$ on the
{\em true} discovery and {\em false} discovery of features by SWA,
respectively. We examine the number of captures of
true features for $k=10$ (full capture), and $k=9+, 8+, \ldots$
for at least $9, 8, \ldots$ true features.
We experiment with $s=5, 8, 10, 15, 20, 30$, and $35$.
Chosen by {Algorithm $2$} in Example 2, $s=30$ is indeed the
best choice based on a simulation study
with $1000$ repetitions. As shown in Table 1, the performance of
true feature detection is improved as the subsample size $s$
increases from 5 to 30. When $s=5$ (or $8$), obviously, detection
rate for more than 5 (or 8) features is zero, because $s<p_0$ with
$q=s$. The two best powers seem to have reached at $s=30$ and $s=35$
in this study. There is little difference between results with $s=30$
and $s=35$, considering that $\beta_1=0.1$ is not very different from
$\beta_1=0$, when $\sigma^2=1$.

\begin{table}[h]
\begin{center}
    \begin{threeparttable}
    \caption{SWA True Feature Detection Comparison at $s=5,8,10,15,20,30$ and $35$}
    {\scriptsize
    \begin{tabular}{crrrrrrr} \hline \hline
        \#Features  & s=5        & s=8       & s=10      & s=15      & s=20      & s=30      & s=35 \\
        Captured        &            &           &           &           &           &           &      \\ \hline
        $10$            & 0\%        & 0\%       & 0\%       & 0\%       & 0\%       & 0\%       & 0.1\% \\
        $9+$            & 0\%        & 0\%       & 0.3\%     & 4.7\%     & 16.8\%    & 26.7\%    & 29\% \\
        $8+$            & 0\%        & 1.2 \%    & 10.2\%    & 46.3\%    & 78.1\%    & 96.2\%    & 96\% \\
        $7+$            & 0\%        & 22.1\%    & 44.0\%    & 86.6\%    & 97.8\%    & 99.9\%    & 100.0\% \\
        $6+$            & 0\%        & 63.7\%    & 80.1\%    & 98.0\%    & 99.9\%    & 100.0\%     &  \\
        $5+$            & 34.4\%     & 90.8\%    & 96.7\%    & 99.8\%    & 100.0\%     &           &  \\
        $4+$            & 63.8\%     & 97.7\%    & 99.9\%    & 100.0\%     &           &           &  \\
        $3+$            & 81.7\%     & 99.6\%    & 100.0\%     &           &           &           &  \\
        $2+$            & 100.0\%      & 100.0\%     &           &           &           &           &  \\
       \hline \hline
    \end{tabular}
    }
    \begin{tablenotes}
      \small
      \item Notes: Entries are percentages of captures of true
        features. ``10'' in the first column means that all 10 true
        feature are captured, while ``$6+$'' means that at least 6
        true features are captured.
    \end{tablenotes}
    \end{threeparttable}
    \end{center}
\label{table:strue}
\end{table}

Table 2 indicates that the false-feature capture rate remains
controlled at $\le5\%$ until $s=30$. At $s=35$, the false discovery
rate is a little bigger than  $5\%$. Hence, $s=30$ is the optimal
value of  $s$  when both the power and false detection rate are
considered.  This outcome  (from Tables 1 and 2)  based on the
repeated experiments  validates our finding by the multipaneling
plot algorithm (MPA).

\begin{table}[h]
\begin{center}
    \begin{threeparttable}
    \caption{SWA False Alarm Comparison at $s=5,8,10,15,20,30$ and $35$}
    {\scriptsize
    \begin{tabular}{crrrrrrr} \hline \hline
    \#Features  & s=5   & s=8   & s=10  & s=15  & s=20  & s=30  & s=35  \\
    Captured        &       &       &       &       &       &       &       \\ \hline
    $ 0$            & 986   & 991   & 987   & 984   & 981   & 956   & 934   \\
    $ 1$            & 14    & 9     & 12    & 15    & 19    & 42    & 58    \\
    $ 2$            &       &       & 1     & 1     &       & 2     & 6     \\
    $ 3$            &       &       &       &       &       &       & 2     \\
    Total           & 1000  & 1000  & 1000  & 1000  & 1000  & 1000  & 1000  \\ \hline \hline
    \end{tabular}
    }
    \begin{tablenotes}
      \small
      \item Notes: Entries are numbers of captures. ``1'' in the
        first column means that exactly one false feature is captured.
    \end{tablenotes}
    \end{threeparttable}
\end{center}
\label{table:sfalse}
\end{table}

\subsection{Comparison with Benchmark Procedures in the Case of
Independent Covariates}
In this section, we compare SWA with 3 benchmark procedures and
Random Forest, when covariates are generated independently as in
model (\ref{eq:simu.model2}) in Example 2. Given by the multipanel
diagnostic plots in Figures 3 and 4, we set $m=5000$ and $s=30$.
We performed a comprehensive, comparative study of SWA with the
Elastic Net, SCAD and MCP as well as Random Forest.

We first compared SWA to the Elastic Net, SCAD and MCP {\em using
the default parameters} in these R packages ``glmnet'' and ``ncvreg''
where their tuning parameter $\lambda$ was chosen corresponding to
the minimum mean cross-validated error for Elastic Net and SCAD,
and using $\sigma=1$ for MCP. The
simulation size is 1000. The False Discovery Rates (FDR)  of
these competing procedures are listed in Table 3. Out of $1000$
independent repetitions, SWA with a Bonferroni control of the
multiplicity at 0.05 FWE performed extremely well, capturing none
of false features in 956 times, exact 1 false feature in 42 of the
trials, exactly 2 nuisance features only twice, and no error in
capturing more than 2 nuisance features.  This SWA had a FDR closest
to  the target level of $0.05$, among all comparison procedures.
With the same Bonferroni
control but adding a stepwise procedure in SWA's last step (Step 4),
the false discovery of the nuisance features by SWA are comparable
to the MCP procedure, both much better than those by the Elastic
Net and SCAD (using their default parameters).
The procedure with a Benjamini-Hochberg (BH) FDR control still did
better than Elastic Net and SCAD (with the default data-dependent
$\lambda$ parameter) procedures in terms of false feature alarm.

\begin{table}[h]
\begin{center}
    \begin{threeparttable}
    \caption{Comparison for False Feature Alarm}
    {\scriptsize
    \begin{tabular}{crrrrrr} \hline \hline
        \#Features      & SWA  w/       & SWA w/    & SWA w/step    & Elastic Net   & SCAD  & MCP   \\
           Captured    & Bonferroni     & BH        & Bonferroni    &               &       &       \\ \hline
            $ 0$        & 956           & 756       & 828           & 7             & 403   & 832   \\
            $ 1$        & 42            & 169       & 142           & 14            & 255   & 151   \\
            $ 2$        & 2             & 53        & 20            & 31            & 175   & 15    \\
            $ 3$        &               & 13        & 6             & 50            & 88    & 2     \\
            $ 4$        &               & 5         & 3             & 65            & 38    &       \\
            $ 5$        &               & 3         & 1             & 76            & 24    &       \\
            $ 6$        &               &           &               & 89            & 10    &       \\
            $ 7$        &               & 1         &               & 91            & 5     &       \\
            $ 8$        &               &           &               & 82            & 2     &       \\
            $ 9$        &               &           &               & 76            &       &       \\
            $\geq10$    &               &           &               & 419           &       &       \\ \hline \hline
    \end{tabular}
    }
    \begin{tablenotes}
      \small
      \item Notes: Entries are numbers of captures. ``1'' in the
        first column means that exactly one false feature is captured.
    \end{tablenotes}
    \end{threeparttable}
    \end{center}
\label{table:false3comp}
\end{table}

Next we compare the competing procedures, SWA, Elastic Net, SCAD and MCP,
in terms of true-feature discovery rates after setting the
regularization parameter $\lambda$ for Elastic Net, SCAD and  $\sigma$
for MCP to match the FDR of SWA at 0.05. The outcome is summarized in Table $4$,
which provides the percentage of times that the true features are captured.
Overall, SWA with a Bonferroni correction (denoted as ``SWA w/ Bonferroni")
under this setting is comparable to SCAD and MCP, while Elastic Net does
not scale up to the performance of either  of SWA, SCAD and MCP.
Specifically, SWA captured all moderate, strong and very
strong features as SCAD and MCP did in the ``7+'' case;  SWA (w/ Bonferroni) had a rate of 96.2\% true-feature
captures in the  ``8+'' case that inclueds a weak signal.  In the ``9+'' case that includes extremely weak features, we do not expect a high capture rate, where SWA (w/ Bonferroni)  had
a true feature capturing rate of 26.7\% which is less than those by SCAD and MCP.
It is possible that if we used a less conservative multiplicity adjustment than
Bonferroni procedure,  such as the Benjamini-Hochberg (BH in short) procedure,
SWA could have captured more very weak features,
though the false discovery may also increase.

In this set of comparisons, the values of $p$ and $p_0$ are typical of
some genetic studies, and are within the implementation limitation of
the SCAD, Elastic Net and MCP procedures. The advantages of SWA are as
follows. First, it does not require a choice of $\lambda$ or determination
of unknown $\sigma$ while providing comparable outcomes for weak to strong
signals. We do need to choose $s$ and $m$, but they are simple to setup.
SWA by default controls the actual FDR to the nominal level, while
penalized  procedures are liberal (as shown in Table 3) by current data
dependent choice of $\lambda$. In the true feature comparison, we adjusted the regularization
parameter of Elastic Net, SCAD and MCP based on the performance
of SWA in this simulation study to equalize the FDR. In a real data application, it is not
clear how to control the FDR to the desired level for each of Elastic Net,
SCAD and MCP individually. Perhaps, we can combine the results of either SWA and
SCAD, or SWA and MCP. See more discussion in Section 7. Third, SWA is
simple. It can handle data with a much larger $p$ than those requiring
all $p$-dimensional data input once for each analysis, as it's a
subsampling procedure. Thus, SWA can scale easily to the cases when the
penalized procedures do not run at the current software implementation.

\begin{table}[h]
\begin{center}
    \begin{threeparttable}
    \caption{Comparison for True Feature Detection}
    {\scriptsize
    \begin{tabular}{crrrrrr} \hline \hline
    \#Features      & SWA w/        & SWA w/    & SWA w/step    & Elastic Net   & SCAD      & MCP       \\
    Captured        &Bonferroni     & BH        & Bonferroni    & FDR-ctrl      & FDR-ctrl  & FDR-ctrl  \\ \hline
    $10$            & 0\%           & 0.6\%     & 0.7\%         & 0\%           & 0.1\%     & 0.1\%     \\
    $9+$            & 26.7\%        & 45.8\%    & 46.8\%        & 0\%           & 41.9\%    & 63.8\%    \\
    $8+$            & 96.2\%        & 98.6\%    & 97.6\%        & 0.8\%         & 99.6\%    & 99.9\%    \\
    $7+$            & 99.9\%        & 100.0\%   & 100.0\%       & 8.1\%         & 100.0\%   & 100.0\%   \\
    $6+$            & 100.0\%       &           &               & 21.4\%        &           &           \\
    $5+$            &               &           &               & 41.5\%        &           &           \\
    $4+$            &               &           &               & 60.2\%        &           &           \\
    $3+$            &               &           &               & 78.9\%        &           &           \\
    $2+$            &               &           &               & 93.6\%        &           &           \\
    $1+$            &               &           &               & 100.0\%       &           &           \\ \hline \hline
    \end{tabular}
    }
    \begin{tablenotes}
      \small
      \item Notes: Entries are percentages of captures. ``10'' in
        the first column means that all 10 true features are captured,
        ``6+'' means that at least 6 true features are captured.
    \end{tablenotes}
    \end{threeparttable}
    \end{center}
\label{table:true3comp}
\end{table}

Random Forest is a machine learning method for classification and
regression. Random Forest and SWA share some similarities such as
random selection of features. However, Random Forest does not
control FDR, but gives a set of importance features for each user
specified number called ``npick,'' the number of features that one
desires to specify in advance. Hence, it is not directly comparable
with SWA,  Elastic Net, SCAD, or MCP in terms of FDR. Regradless,
we evaluated Random Forest performance on the same data from Example 2.
In comparison to SWA, Table 5 summarizes the percentage of true
features among top ``npick'' features found by Random Forest measure
of importance. When ``ntree,'' another user specified number in
Random Forest calculation, is set to be ntree $=1000$, the resulting
true-feature capture percentage improves as ``npick'' increases from
10 to 30, which would mean that the FDR would increase at least to
10 (50\%) for npick $=20$, and 20 (67\%) for npick $=30$. When npick
$=10$, the number of true features, only strong and moderate features
are captured well by Random Forest. With npick $=10 \& 20$, we also
increased the number of trees, ``ntree,'' from 1000 to 2000, which did
not help much.  In comparison, SWA performs well for all levels,
controlling the FDR and providing good true feature detection rates.

\begin{table}[h]
\begin{center}
    \begin{threeparttable}
    \caption{Random Forest True Feature Detection Comparison}
    {\scriptsize
    \begin{tabular}{crrrrrr} \hline \hline
        \#Features      & npick=30  & npick=20  & npick=10  & npick=20  & npick=10   & SWA w/\\
        Captured        & ntree=1K  & ntree=1K  & ntree=1K  & ntree=2K  & ntree=2K   &Bonferroni\\ \hline
        $10$            & 1.3\%     & 0.1\%     & 0\%       & 0.1\%     & 0\%        & 0\%    \\
        $9+$            & 7.9\%     & 1.9\%     & 0\%       & 2.7\%     & 0.1\%      & 26.7\% \\
        $8+$            & 30.4\%    & 13.7 \%   & 0.7\%     & 17.3\%    & 1.7\%      & 96.2\% \\
        $7+$            & 67.6\%    & 44.9\%    & 13.6\%    & 52.2\%    & 13.1\%     & 99.9\% \\
        $6+$            & 93.1\%    & 81.1\%    & 46.3\%    & 82.2\%    & 51.5\%     & 100.0\%  \\
        $5+$            & 99.2\%    & 97.2\%    & 98.7\%    & 97.7\%    & 86.8\%     &      \\
        $4+$            & 100.0\%     & 99.9\%    & 100.0\%     & 100.0\%     & 99.4\%     &      \\
        $3+$            &           & 100.0\%     &           &           & 100.0\%      &      \\
         \hline \hline
    \end{tabular}
    }
    \begin{tablenotes}
      \small
      \item Notes: Entries are percentages of captures.
        ``10'' in the first column means that all 10 true
        feature are captured, ``$6+$'' means that at least 6
        true features are captured.
    \end{tablenotes}
    \end{threeparttable}
    \end{center}
\label{table:rforest}
\end{table}

Based on the above outcome, we shall only compare SWA with Elastic Net,
SCAD, and MCP in the correlated and ultra-high dimension cases below.

\subsection{Comparison with Correlated Covariates}
We repeat our comparison when the covariates are correlated as it
used in \citet{tib}.  Specially, in Example 2, let correlation
between $x_i$ and $x_j$ be $\rho^{|i-j|}$ for $i,j=1,\cdots,11$
with $\rho =0.5$, and $\sigma =1$ and $=3$ respectively. The
results for case 1 of $\sigma =1$ are  given in Tables $6 \& 7$,
while those for case 2 of $\sigma =3$ are  in Tables $8 \& 9$.

For Case 1, in terms of false feature alarm, SWA and SCAD made
the fewest mis-detection, which is followed by MCP and then Elastic
Net. In terms of true feature detection, after adjusting benchmark
procedures to match the same FDR as that of SWA, SWA  is found to
be comparable with the benchmark procedures as shown in Table 7,
with MCP to be the best performer, slightly.

\begin{table}[h]
\begin{center}
    \begin{threeparttable}
    \caption{False Feature Alarm in Correlated Case 1:
            $\rho =0.5$ and $\sigma =1$}
    {\scriptsize
    \begin{tabular}{crrrrr} \hline \hline
        \# features     & SWA w/        & SWA w/        & Elastic Net   & SCAD  & MCP   \\
            captured    & Bonferroni    & BH            &               &       &       \\ \hline
            $ 0$        & 961           & 768           & 264           & 977   & 822   \\
            $ 1$        & 36            & 171           & 195           & 21    & 163   \\
            $ 2$        & 3             & 42            & 143           & 2     & 14    \\
            $ 3$        &               & 15            & 129           &       & 2     \\
            $ 4$        &               & 2             & 61            &       &       \\
            $ 5$        &               & 2             & 68            &       &       \\
            $ 6$        &               &               & 37            &       &       \\
            $ 7$        &               &               & 28            &       &       \\
            $ 8$        &               &               & 17            &       &       \\
            $ 9$        &               &               & 21            &       &       \\
            $\geq10$    &               &               & 37            &       &       \\ \hline \hline
    \end{tabular}
    }
    \begin{tablenotes}
      \small
      \item ``1'' in the first column means that
            exactly one false feature is captured.
    \end{tablenotes}
    \end{threeparttable}
    \end{center}
\label{tab:falsecorr1}
\end{table}

For Case 2, simulation variation is increased to $\sigma =3$. As
shown in Table 8, SWA again has the fewest mis-detection. In terms
of true feature detection, after matching benchmark procedures'
FDR at the same rate of SWA, SWA is found to be comparable with
the competing procedures for detecting the moderate and stronger
features, while Elastic Net does better in detecting the weak
features.

\begin{table}[hbt]
\begin{center}
    \begin{threeparttable}
    \caption{True Feature Detection in Correlated Case 1:
            $\rho =0.5$ and $\sigma =1$}
    {\scriptsize
    \begin{tabular}{crrrrr} \hline \hline
    \# of features  & SWA w/        & SWA w/        & Elastic Net     & SCAD        & MCP       \\
    captured        & Bonferroni    & BH            & FDR-ctrl        & FDR-ctrl    & FDR-ctrl  \\ \hline
    $10$            & 0\%           & 0\%           & 2.5\%           & 1.9\%       & 0.7\%     \\
    $9+$            & 21.5\%        & 45.7\%        & 50.1\%          & 60.2\%      & 50.7\%    \\
    $8+$            & 94.4\%        & 98.8\%        & 99.8\%          & 99.2\%      & 100\%     \\
    $7+$            & 100\%         & 100\%         & 100\%           & 100\%       &           \\
     \hline \hline
    \end{tabular}
    }
    \begin{tablenotes}
      \small
      \item Notes: Entries are percentages of captures. ``10'' in
        the first column means that all 10 true features are captured,
        ``6+'' means that at least 6 true features are captured.
    \end{tablenotes}
    \end{threeparttable}
    \end{center}
\label{table:truecorr1}
\end{table}

\begin{table}[h]
\begin{center}
    \begin{threeparttable}
    \caption{False Feature Alarm in Correlated Case 2:
            $\rho =0.5$ and $\sigma =3$}
    {\scriptsize
    \begin{tabular}{crrrrr} \hline \hline
        \# features     & SWA w/        & SWA w/        & Elastic Net   & SCAD  & MCP   \\
            captured    & Bonferroni    & BH            &               &       &       \\ \hline
            $ 0$        & 875           & 567           & 299           & 73    & 638   \\
            $ 1$        & 10            & 234           & 213           & 99    & 272   \\
            $ 2$        & 21            & 101           & 143           & 142   & 73    \\
            $ 3$        &               & 57            & 82            & 127   & 13    \\
            $ 4$        &               & 19            & 69            & 117   & 3     \\
            $ 5$        &               & 13            & 44            & 112   & 1     \\
            $ 6$        &               & 6             & 31            & 95    &       \\
            $ 7$        &               & 2             & 21            & 68    &       \\
            $ 8$        &               & 1             & 20            & 68    &       \\
            $ 9$        &               &               & 14            & 42    &       \\
            $\geq10$    &               &               & 55            & 57    &       \\ \hline \hline
    \end{tabular}
    }
    \begin{tablenotes}
      \small
      \item Notes: ``1'' in the first
        column means that exactly one true feature is captured.
    \end{tablenotes}
    \end{threeparttable}
    \end{center}
\label{table:falsecorr2}
\end{table}

\begin{table}[h]
\begin{center}
    \begin{threeparttable}
    \caption{True Feature Detection from Correlated Case 2:
            $\rho =0.5$ and $\sigma =3$}
    {\scriptsize
    \begin{tabular}{crrrrr} \hline \hline
    \# features     & SWA w/        & SWA w/        & Elastic Net     & SCAD        & MCP       \\
    captured        & Bonferroni    & BH            & FDR-ctrl        & FDR-ctrl    & FDR-ctrl  \\ \hline
    $10$            & 0\%           & 0\%           & 0.8\%           & 0.1\%       & 0\%       \\
    $9+$            & 0.1\%         & 0.5\%         & 7.8\%           & 2.8\%       & 0.3\%     \\
    $8+$            & 1.7\%         & 8.9\%         & 35.5\%          & 23.5\%      & 10.5\%    \\
    $7+$            & 20.3\%        & 51\%          & 80.6\%          & 59.7\%      & 45.2\%    \\
    $6+$            & 71.1\%        & 92.2\%        & 98.0\%          & 86.6\%      & 86.1\%    \\
    $5+$            & 95.5\%        & 99\%          & 99.9\%          & 97.6\%      & 98.5\%    \\
    $4+$            & 99.8\%        & 99.8\%        & 100\%           & 100\%       & 100\%     \\
    $3+$            & 100\%         & 100\%         &                 &             &           \\
    \hline \hline
    \end{tabular}
    }
    \begin{tablenotes}
      \small
      \item Notes: Entries are percentages of captures. ``10'' in
        the first column  means that all 10 true features are captured,
        ``6+'' means that at least 6 true features are captured.
    \end{tablenotes}
    \end{threeparttable}
    \end{center}
\label{table:truecorr2}
\end{table}

\subsection{Comparison in Ultra high $p=1000$ Dimensions}
Consider the model in Example 2 with the same $p_0=10$ and $n=80$,
but let $p =1000$ instead of $p=100$. SWA did not perform well without
a pre-screening step. Since it is common to perform a pre-independence
screening procedure [\citet{fan2}] for variable selection,
we added a pre-screening step by selecting $100$ $x$-variables
with the strongest marginal correlation to the response variable
$y$. SWA procedure with a repetition size $m = 5000$ and $s=30$
was applied afterwards. SWA (w/ Bonferroni adjustment of $100$ or $30$)
outperformed, Elastic Net, SCAD and
MCP using default parameters, in false feature detection (in
Table 10). After setting the smoothing parameters in Elastic Net,
SCAD and MCP to match the FDA at the same level, SWA was found still
competitive overall as shown in Table 11. In this ultra high dimensional
case, we recommend using the Bonferroni adjustment at the rate of
$100$, the size that was adjusted to after a pre-screening step.
Arguably,  using a Bonferroni adjustment at the rate of $30$, the size
of subsample size  $s$ and the size of the semi-finalists $q$, is also
reasonable.

\begin{table}[htb]
\begin{center}
    \begin{threeparttable}
    \caption{False Feature Detection in Ultra High $p=1000$ Case}
    {\scriptsize
    \begin{tabular}{crrrrr} \hline \hline
        \# features     & SWA           & SWA adjusted  & Elastic Net   & SCAD  & MCP   \\
            captured    & $(0.05/100)$  & $(0.05/30)$   &               &       &       \\ \hline
            $ 0$        & 965           & 913           &               & 136   & 714   \\
            $ 1$        & 29            & 71            &               & 147   & 249   \\
            $ 2$        & 6             & 14            &               & 131   & 35    \\
            $ 3$        &               & 2             &               & 128   & 2     \\
            $ 4$        &               &               &               & 87    &       \\
            $ 5$        &               &               &               & 77    &       \\
            $ 6$        &               &               &               & 65    &       \\
            $ 7$        &               &               & 1             & 60    &       \\
            $ 8$        &               &               &               & 48    &       \\
            $ 9$        &               &               & 1             & 22    &       \\
            $\geq10$    &               &               & 998           & 29    &       \\ \hline \hline
    \end{tabular}
    }
    \begin{tablenotes}
      \small
      \item Notes: Entries are numbers of true feature captures.
        ``1'' in the first column means that exactly one feature is captured.
    \end{tablenotes}
    \end{threeparttable}
    \end{center}
\label{table:falselargep}
\end{table}

\begin{table}[h]
\begin{center}
    \begin{threeparttable}
    \caption{True Feature Detection in Ultra High $p=1000$ Case}
    {\scriptsize
    \begin{tabular}{crrrrr} \hline \hline
    \# features     & SWA adjusted  & SWA adjusted  & Elastic Net   & SCAD      & MCP       \\
    captured        & $(0.05/100)$  & $(0.05/30)$   & FDR-ctrl      & FDR-ctrl  & FDR-ctrl  \\ \hline
    $10$            & 0\%           & 0\%           &               &           &           \\
    $9+$            & 27.9\%        & 38.1\%        &               & 3.1\%     & 31.4\%    \\
    $8+$            & 96.6\%        & 98\%          &               & 58.4\%    & 99.7\%    \\
    $7+$            & 100\%         & 100\%         &               & 96.5\%    & 100\%     \\
    $6+$            &               &               & 0.3\%         & 99.7\%    &           \\
    $5+$            &               &               & 5.9\%         & 100\%     &           \\
    $4+$            &               &               & 19.1\%        &           &           \\
    $3+$            &               &               & 33.2\%        &           &           \\
    $2+$            &               &               & 41.5\%        &           &           \\
    $1+$            &               &               & 100\%         &           &           \\ \hline \hline
    \end{tabular}
    }
    \begin{tablenotes}
      \small
      \item Notes: Entries are percentages of captures. ``10'' in
        column 1 means that all 10 true features are captured, ``6+''
        means that at least 6 true features are captured.
    \end{tablenotes}
    \end{threeparttable}
    \end{center}
\label{table:truelargep}
\end{table}

\section{Application and Double Assurance Procedure}

In this section, we apply our SWA to the ovarian cancer data.
There are $n=561$ samples, each having $p=12042$
mRNA gene expressions. This data set neither contains an obvious
response variable nor has control cases.  We hence choose the
Cyclin E gene, termed as ``CCNE1'', as the response in this case
study. ``CCNE1'' is a good surrogate for the survival, according
to the cBioPortal for Cancer Genomics (http://cbioportal.org) by
\citet{gao}, 25\% of the 591 ovarian cancer patients were found
to have their ``CCNE1'' gene altered, and the subjects with the
alteration of ``CCNE1'' gene had a significantly lower overall
survival rate (P-value$=1.619\times 10^{-4}$) as shown in the
right side of Figure 5. 
Hence, we are interested in finding mRNAs that
collectively are significant explanatory variables or features
for ``CCNE1.'' Finding these interesting explanatory variables is
helpful to examine/find pathways, not just one gene, responsible
for survival.

\begin{figure}[htb]
\begin{minipage}[t]{0.5\textwidth}
    \includegraphics[width=1.1\textwidth]{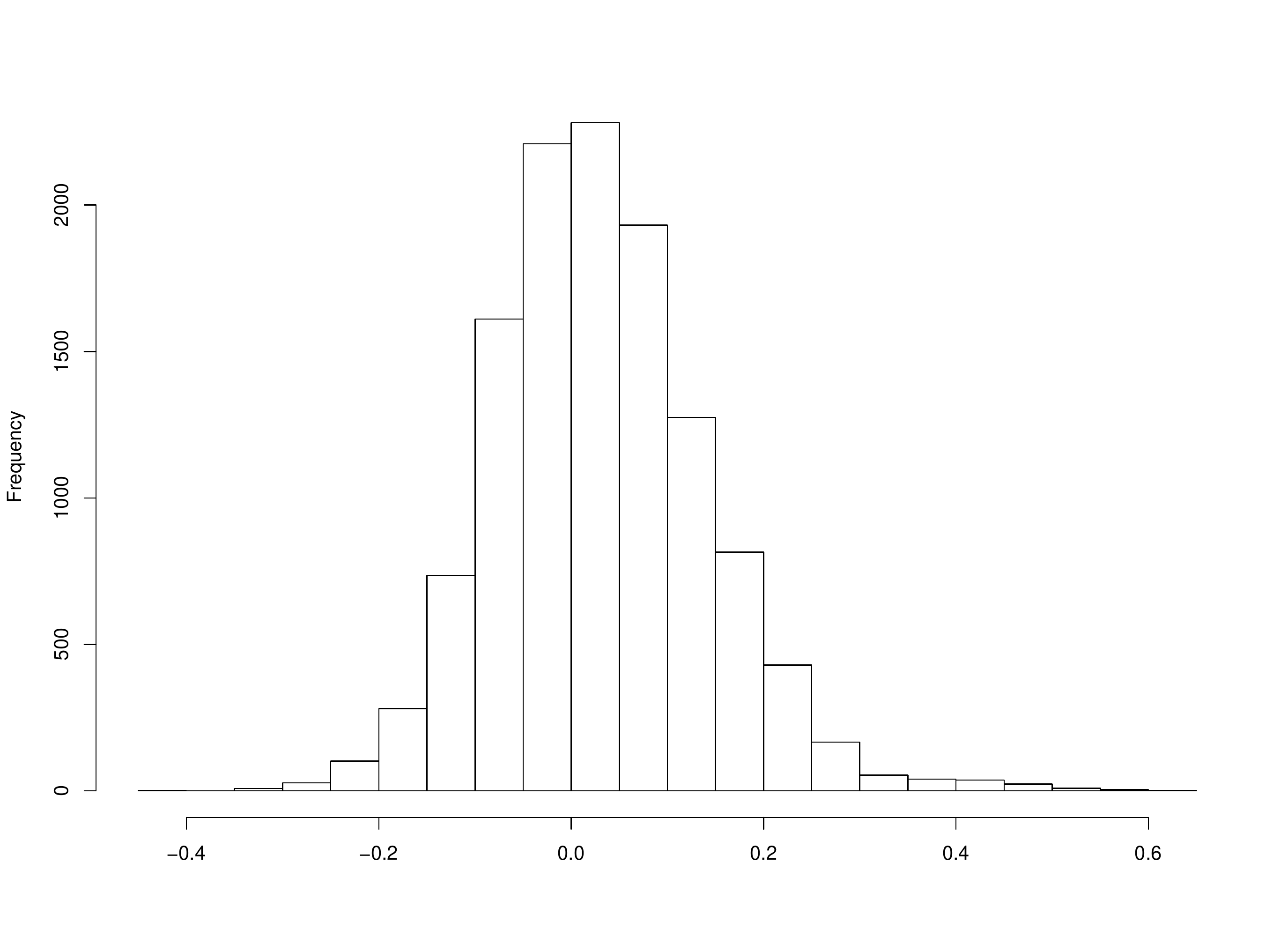}
\end{minipage}
\begin{minipage}[t]{0.5\textwidth}
    \includegraphics[width=1.2\textwidth]{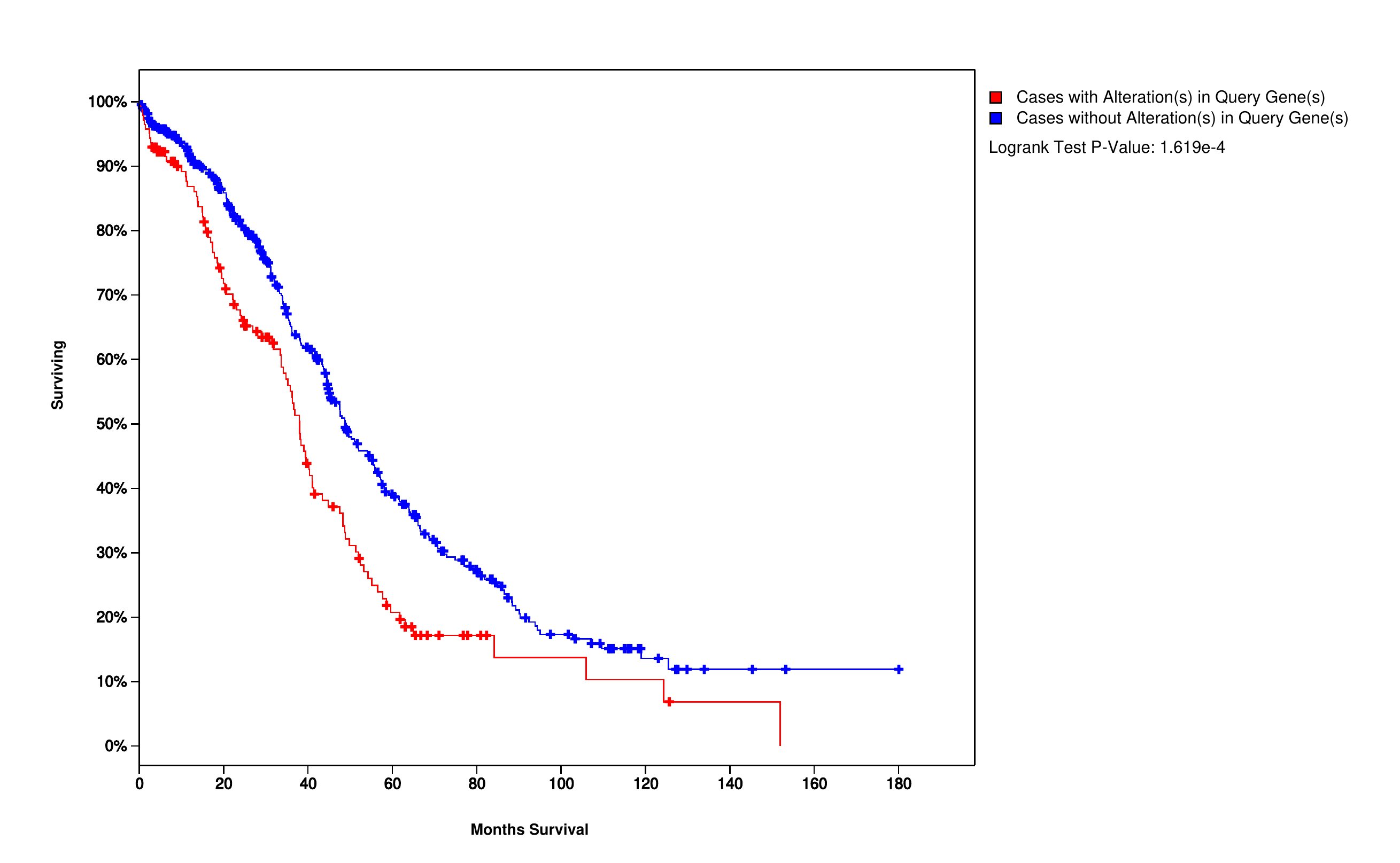}
\end{minipage}
    {\em Figure 5: Left: Histogram of correlation coefficients. Right: Kaplan-Meier plot for the ovarian cancer data using cBioPortal}
\end{figure}

First, we activate a screening step based on the distribution of sample
correlation in the left side of Figure 5, 
to reduce the number of features from $412024$ to $p=901$ by selecting
the features with an absolute value of correlation $\geq 0.20$.
Then, we apply SWA with a repetition size $m=10,000$ and subsampling
sizes $s$ from $S = \{5, 8, 10, 15, 20$ and $30 \}$ to choose the
`optimal' subsample $s_0$. Guided by {Algorithm $2$}, examining
the fixed scale multipanel plot in Figure 6, 
we find that significant changes in magnitude occur at $s=20$. Combining
this fixed-scale indication with the free-scale information in
Figure 7, 
we observe that $s=20$ and $30$ lead to plots with a stable `elbow' point
and a stable upper arm set. Thus, the choice of $s_0=min(s_i)=20$ is the
optimal subsample size that is stabilized in both scale and shape. However,
as an enhancement to the practice, we propose one more step to implement a
`Double-Assurance Procedure' below.

\vspace{0.1in}
{\bf Algorithm 3: Double Assurance Procedure}
\begin{enumerate}
    \item Consider $SWA(s,m,q)$ for $s=s_0$ and a few $s_i\in S$ with $s_i<s_0$ that
        seemed to  have led to semi-stable points.
    \item Combine the final significant features in  ${\rm SWA}(s_0,m,q)$ and
        ${\rm SWA}(s_i,m,q)$ for these $s_i$'s (here, $s_i=10$ and $15$)
        into a combined prediction set $\bf{x'}$, and then run another base
        analysis of $\bf{h}$ for $\bf{y\sim x'}$, and perform a variable
        selection step on the outcomes.
\end{enumerate}


\begin{center}
    \includegraphics[width=0.97\textwidth]{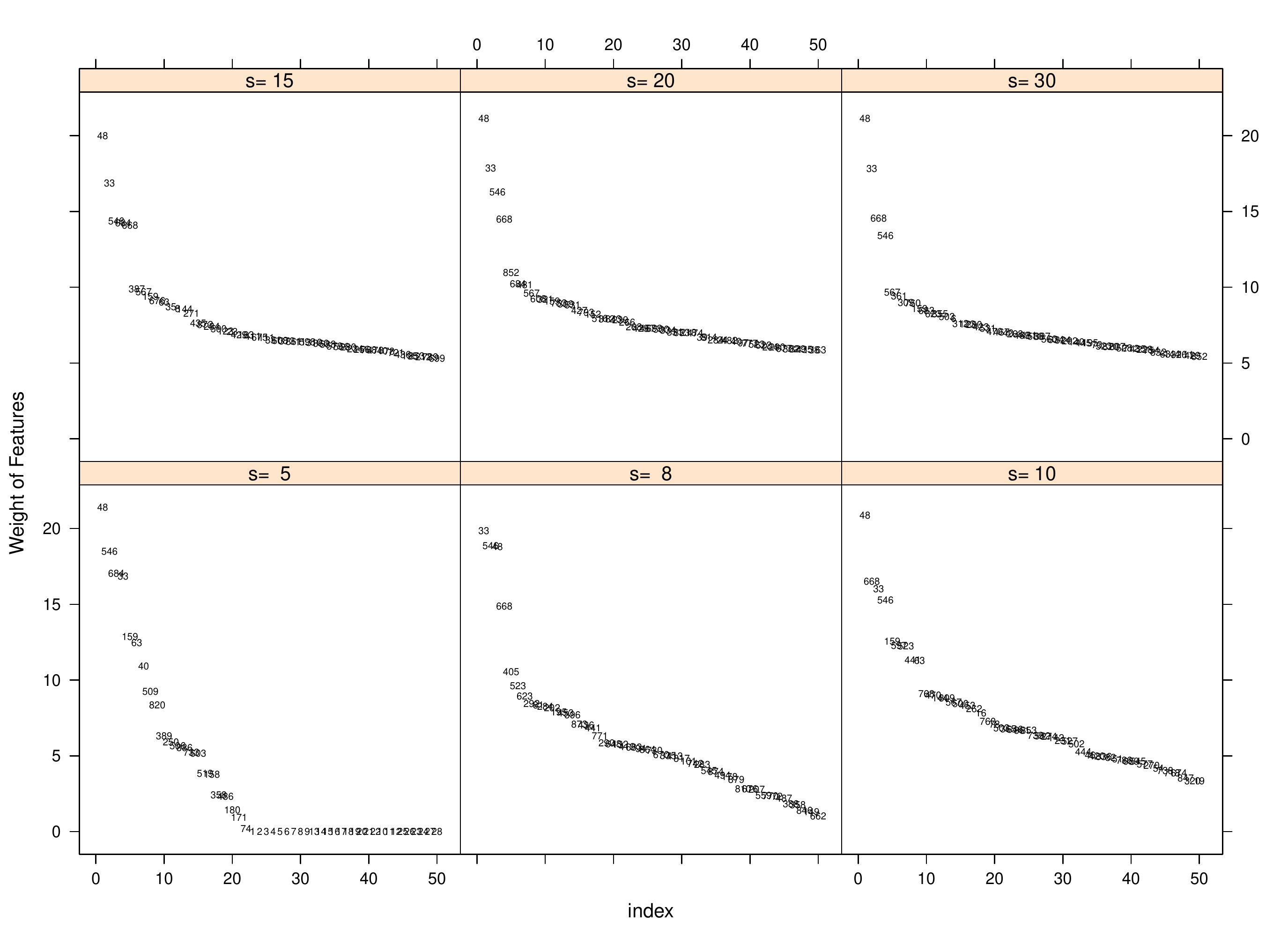}\\
  {\em Figure 6: Fixed-scale multi-panel diagnostics plot for ovarian cancer study.}\\
    \vspace*{\floatsep}
    \includegraphics[width=0.97\textwidth]{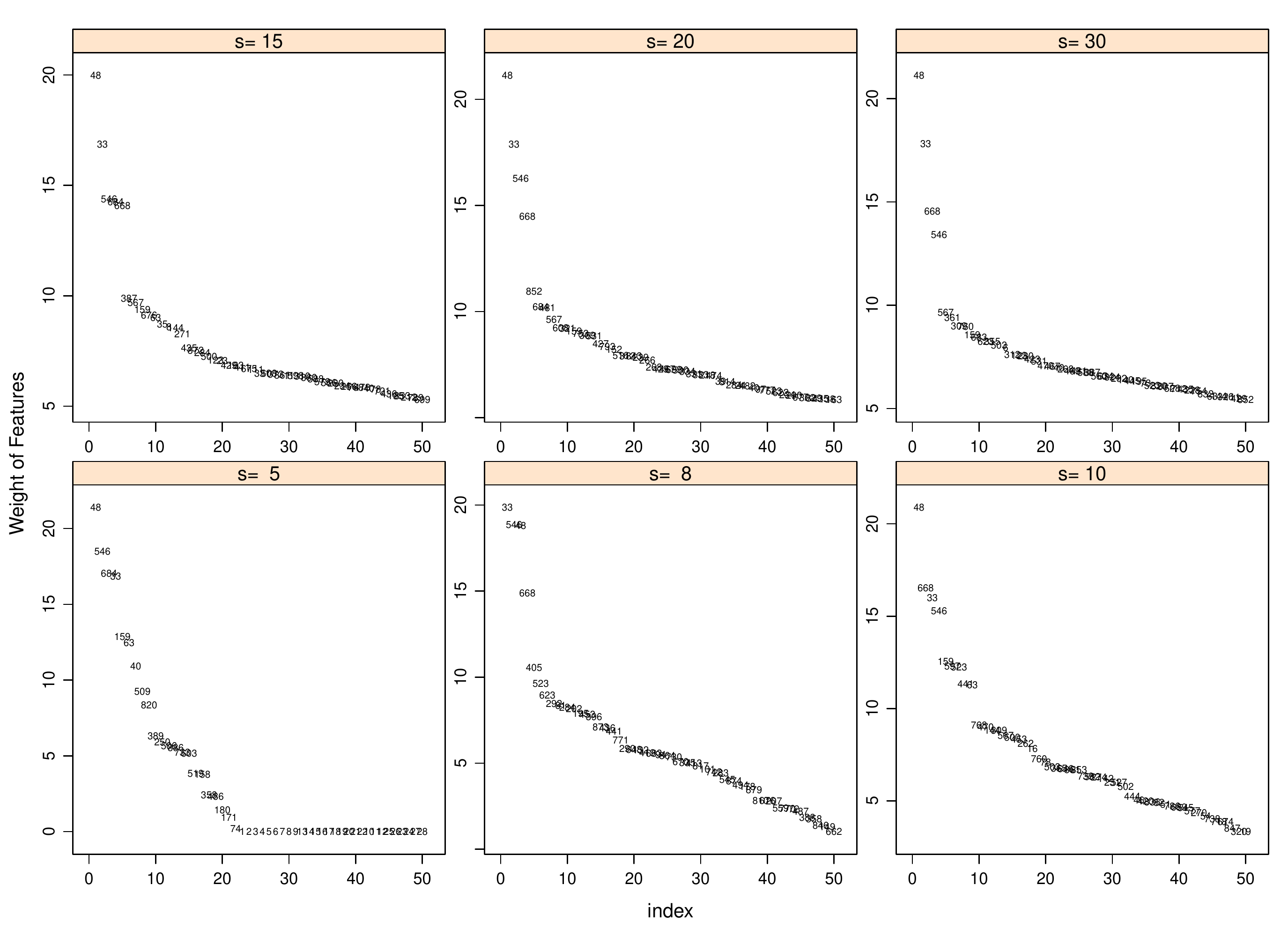}\\
 {\em Figure 7: Free-scale multi-panel diagnostics plot for ovarian cancer study.}
\end{center}

The final significant features from {\bf Algorithm 3} are listed in
Table 12. Our set of features is sensible based on a STRING analysis
of our detected features using functional protein association networks
(http://string-db.org/) by \citet{string}. The STRING analysis shows
that our discovered features contain functionally important genes;
in particular ``POLQ'' and ``RAD51'' form a special pathway in which
``POLQ'' blocks ``RAD51''-mediated recombination. Together,
``POLQ'' and ``RAD51''
correlate with defects in homologous recombination (HR) repair
published in {\it Nature} [\citet{nature}]. In addition, pathway
analysis shows that the network has significantly more interactions
than expected (P-value$<0.004$). Both our findings and their results
reveal a synthetic lethal relationship between the HR pathway and
``POLQ''-mediated repair in Ovarian Cancer. The biological application
of the discovery of features in Table 12 may go beyond the statements
above. In a separate biological manuscript, we apply another new
bioinformatics tool to a combined set of these features with other
``BRCA'' genes to explore potentials for drug target for ovarian cancer.


\begin{table}[h]
\begin{center}
    \begin{threeparttable}
    \caption{Final Results from Double-Assurance Procedure}
    {\scriptsize
    \begin{tabular}{clrl} \hline \hline
    \# of features  & Gene Name     & Estimate      & P-value       \\ \hline
    $1$             & CDKN2A        & 0.26554       & $<$ 2e-16     \\
    $2$             & C19orf2       & 0.28228       & 4.94e-10      \\
    $3$             & PLEKHF1       & 0.21810       & 5.08e-09      \\
    $4$             & POLQ          & 0.35074       & 4.25e-06      \\
    $5$             & TMEM30B       & -0.13503      & 8.61e-06      \\
    $6$             & GLCE          & -0.16539      & 5.93e-05      \\
    $7$             & POP4          & 0.20944       & 7.06e-05      \\
    $8$             & GIPC1         & 0.19072       & 0.000393      \\
    $9$             & C17orf53      & 0.43510       & 0.000529      \\
    $10$            & C15orf15      & -0.15195      & 0.000586      \\
    $11$            & TM2D3         & -0.14747      & 0.000712      \\
    $12$            & CRIM1         & -0.14717      & 0.000768      \\
    $13$            & RAD51         & 0.27486       & 0.002840      \\ \hline \hline
    \end{tabular}
    }
    \end{threeparttable}
    \end{center}
\label{table:double-dip}
\end{table}

\section{Discussion and Conclusions}

We have provided a new procedure, Subsampling Winner Algorithm
(SWA), for finding important features in large-$p$ regression. We
performed a comprehensive study to compare SWA with the benchmark
procedures, LASSO, Elastic Net, SCAD, MCP and Random Forest. We
also applied the SWA to analyze  the genomic data on ovarian cancer.
Our study revealed a meaningful ovarian cancer pathway verified
using the STRING analysis, as well as new genes that have become
candidates to be examined biologically.

Our subsampling approach  represents a paradigm shift from
the popular approach using a penalized criterion in selecting
important/true features from  $p$ covariates when $p\gg n$.
The penalized procedures approximate the solution to {\em a
penalized criterion based on entire sample}. The SWA is
{\em an ensemble of simple procedures $h$ performed on subsamples},
by lifting the good performance of $h$ in low $s$-dimension
($\ll n$) to the high $p$-dimension ($\gg n$) using our scoring
function strategically.  SWA is analogous  to  the method for
selecting  national merit scholars.

The advantages of SWA are given below.

First, the SWA is virtually dimensionless, due to its subsampling
nature and the simplicity of its base procedure (without a need
for finding a penalizing or turning parameter).  Hence, it can
scale  easily  to an enormously large $p$, especially when $p$
is too large to use a benchmark, penalized procedure that requires
data inputted in an existing software in one pass. Also due to the
simplicity and subsampling nature, SWA can be made ``embarrassingly
parallel" and distributed over multiple cores for fast computation
and  for enlarging  data. We leave this kind of coding refinement
for multiple cores or high performance computers to the future.

Second,   when the large p is manageable, SWA is shown to be
competitive to benchmark procedure in our simulation studies.
In terms of the false discovery rate (FDR), we compared  the
SWA using a simple Bonferroni multiplicity adjustment (denoted
as SWA w/Bonferroni) with LASSO, Elastic Net, SCAD, MCP and
Random Forest (RF). We used the default parameter $\lambda$ based
on a cross-validated mean square errors  for Elastic Net and SCAD,
and the known $\sigma=1$ for MCP,  as well as various parameter
choices for RF. The SWA {\em controlled the false discovery rate
(FDR) closest}  to the target value (Table 3, 6, 8 and 10).
Random Forest does not explicitly control FDR and hence is not
comparable (Table 5). The advantage of Random Forest is  not at
controlling  FDR, but at its prediction precision and perhaps at
providing a set of  importance features as a first-stage dimension
reduction candidates.


In terms of true discovery rate (TDR),  we compared the true feature selection  after equalizing the
FDRs for all penalized procedures to that of the SWA.  The good performers are summarized in Table 13 based on the comprehensive results in
Tables 4,7,9,11 and 12. Specifically, in the case of
{\em independent covariates}, SWA, SCAD and MCP were
comparable except for the very weak features, where MCP was the
(slight) winner followed closely by SCAD and SWA (Table 4).
Elastic Net did not perform well for the independent
covariates case (Table 4), but was comparable with  SWA, SCAD and MCP in Cases 1 and 2 of  the {\em correlated-covariates} (Tables 7 and 9), and
performed the best slightly  for the correlated case 2 (Table 9).
SWA and MCP are clear winners in the {\em ultra-high dimension
case}   with independent covariates (Table 11).  In the ultra high dimension case, it is important to add a prescreening procedure as we and all penalized procedures have done.

\begin{table}[h]
\begin{center}
    \begin{threeparttable}
    \caption{Results Summary and Recommendations for Large, Manageable p}
    {\scriptsize
    \begin{tabular}{lll|l} \hline \hline
    \multicolumn{1}{l}{Cases}             & Good Performers       & Good Performers           & \multicolumn{1}{c}{Practical}         \\
                        & based on FDR        & based on TDR             & \multicolumn{1}{c}{Recommendation}\\ \hline
    Independent         & SWA (Table 3)         & MCP, SCAD \&               & \multirow{3}{*}{
        \begin{tabular}[c]{@{}l@{}} {\bf A}. SWA with an added \\ double assurance\\  (Algorithm 3); \\  \multicolumn{1}{c}{or}\\
        {\bf B.} Combination of SWA \\ \& a penalized procedure\\ (Algorithm 4) \end{tabular}} \\
                        &                       & SWA (Table 4)             &           \\
    Correlated Case1    & SCAD \& SWA (Table 6)   & MCP, SCAD, ElasticNet        &           \\
                        &                       & \& SWA (Table 7)        &           \\
    Correlated Case2    & SWA (Table 8)         & ElasticNet, SWA,         &           \\
                        &                       & MCP \& SCAD (Table 9)       &           \\
    Ultra High          & SWA (Table 10)        & MCP, \& SWA (Table 11)       &           \\ \hline \hline
    \end{tabular}
    }
    \begin{tablenotes}
      \small
      \item {\em \small Notes:  Entries in Columns 2 and 3 are good performers based on the results from the table cited. Recommendations  in Column 4 were made based on  both FDR and TDR. }
    \end{tablenotes}
    \end{threeparttable}
    \end{center}
\label{table:summary}
\end{table}

In all these comparisons,  for Elastic Net, SCAD and MCP we used a trial-and-error method to modify
the $\lambda$ and $\sigma$ from their default or initial values
 to match the FDR at 0.05.  In practice, without knowing
the truth, this trial-and-error adjustment  to match actual 0.05 rate is
impractical. In practice,  thus, we recommend using SWA with a double assurance as in Algorithm 3,  or a combination of
SWA with  one of good performers from Elastic Net, SCAD, and MCP
as shown in Table 13 by the following Algorithm 4.

{\em Algorithm 4}: Combine features from
SWA and a good performing penalized procedure and run our base procedure $h$ on this combined predictor set to lead to the final set of
features.

We do not have a rigorous mathematical statement or proof of SWA's optimality, although we gave a rationale for its obvious consistency if given the SOIL condition, as \citet{ye} did.  SWA's excellent performance  and scalability to
an ultra-large $p$  may bear some
similarity to that of the deep learning procedure [\citet{LeCon}] as follows.
A deep learning algorithm is built on many layers of linear combination of
{\em simple} basis functions, while SWA is based on
a fusion of many {\em simple} analyses of subsamples.
   The
purpose of  deep learning parallels to that of RandomForest, being excellent
for prediction, while SWA aims directly at feature selection.
Another reason for the good performance is perhaps at SWA's selection of
important features from an ensemble of (subsample) models  in
the same spirit of the SOIL conditions of \citet{ye}.  The difference between our SWA
and the ensemble of \citet{ye} is that we extract final
useful set of features from {\em simple analyses of subsamples} (hence, SWA is scalable to very high dimensions) while
\citet{ye}'s ensemble obtains final results from {\em penalized or other
iterated procedures of whole samples}.

Although SWA does not need to choose a penalizing
parameter $\lambda$ or require an estimate of $\sigma$ in
advance, SWA does need to choose a subsample size $s$
and the number of subsample iterations $m$; fortunately,
they are straightforward to select and our selection
automatically controls the FDR to the target level.
Specifically, the SWA is reasonably stable as long as
$p_0\leq s\leq 3p_0$ with a sufficiently large $m$; and
$s$ can be chosen more specifically by our simple
multi-paneling plot with a fixed $m=5000$.  Another
parameter $q$,  the  number of ``semifinalists" in SWA
is set to be $s$ in our simulation experiments herein
and was shown to be satisfactory. However, this $q$ can
easily be enlarged to  $1.5s$ or $2s$ to be conservative.
We also proposed a double-enhancement procedure, which
is similar to enlarging $q$ from $s$,  for practical
applications in Section 6.

The SWA in this paper is developed for linear regression,
the most fundamental model. The SWA can be generalized
for a generalized linear model, or to a nonparametric
classification function, as suitable for other type of data
and analysis objectives. The key is to choose $h$ properly
and generalize the scoring function $w$ accordingly. See
SWA for principal component analysis [\citet{liu2}] used
different $h$ and $w$. We leave the study for these
generalized models using SWA to the future.

It is also worth to note the following caveats before using any new
advanced procedure to analyze a data set. First, preprocessing
data is essential as it can remove a large number of obviously
nuisance features cheaply and quickly in advance. Second, an
ensemble of statistical procedures is commonly needed for
analyzing a large and complex data, for example,  the ensemble
used for analyzing movies  (a sequence of images) in a clinical
study from \citet{wangx}  and \citet{bogie} includes both image
preprocessing and statistical analysis procedures. Third, correct
modeling is critical to features selection for examining the effect
of possible influencing factors. Under different models, the effects
of various features may be shown differently.

Finally, we advocate that any biological implications be verified
by independent biological experiments or another instrument.
We used cBiportal and STRING analysis tools to our findings by SWA.
This is consistent to the recommendation made by
\citet{Wasserstein}.
We are quite pleased in discovering that our study of the ovarian cancer
data has led to important pathways, and also that there is a potential
for biological applications beyond what is presented in Section 6.

\appendix

\section{Regularity Conditions and Proofs}\label{app}

\begin{proof}
The idea is to first calculate the probability that all the important
features are selected and detected in one subsample, then choose the required
$m$ such that this probability is at least $1-\gamma$.

Since $s\ge p_0$, there exists a chance that all the important features are
selected in a subsample so that the important features can be detected at once.
For each subsample, the probability that all $p_0$ important features are selected
is
\begin{equation}
    \alpha=\frac{\binom{p-p_0}{s-p_0}}{\binom{p}{s}}=\frac{\frac{(p-p_0)!}{(s-p_0)!(p-s)!}}{\frac{p!}{s!(p-s)!}}
            =\frac{(p-p_0)!s!}{(s-p_0)!p!}=\frac{s\dotsb(s-p_0+1)}{p\dotsb(p-p_0+1)}
\end{equation}
It is known that $s<p$, we can obtain upper and lower bounds for $\alpha$ based
on the above expression
\begin{equation}
    \left(\frac{s-p_0+1}{p-p_0+1}\right)^{p_0} \le \alpha \le \left(\frac{s}{p}\right)^{p_0}.
\label{eq:msize.02}
\end{equation}
Assume the random subsamples are independent of each other, the probability
that all $p_0$ important features are selected at least once in one subsample
after m trials is $1-(1-\alpha)^m$ Setting it to be
\begin{equation}
    1-(1-\alpha)^m=1-\gamma,
\end{equation}
we have
\begin{equation}
    m=\frac{log\gamma}{log(1-\alpha)}
\label{eq:msize.03}
\end{equation}
and can thus obtain the upper and lower bounds of $m$ from the upper and
lower bounds of $\alpha$, which are the ones given in (\ref{eq:msize.01}).
\end{proof}



\end{document}